\documentclass[sigconf, screen]{acmart}

\AtBeginDocument{%
  \providecommand\BibTeX{{%
    \normalfont B\kern-0.5em{\scshape i\kern-0.25em b}\kern-0.8em\TeX}}}

\setcopyright{acmcopyright}
\copyrightyear{2022}
\acmYear{2022}
\acmDOI{XXXXXXX.XXXXXXX}

\acmConference[Conference acronym 'XX]{enter conference title from your rights confirmation emai}{June 03--05,
  2018}{Woodstock, NY}
\acmPrice{15.00}
\acmISBN{978-1-4503-XXXX-X/18/06}

\usepackage{booktabs}       %
\usepackage{amsfonts}       %
\usepackage{amsmath}
\usepackage{amsthm}
\usepackage{mathtools}
\usepackage{nicefrac}       %
\usepackage{microtype}      %
\usepackage{enumitem}
\usepackage{dsfont}
\usepackage{hyperref}
\usepackage{siunitx}
\usepackage{bm}
\usepackage{xcolor}
\usepackage{tcolorbox}
\usepackage{mleftright}
\usepackage{stmaryrd}
\usepackage{numprint}
\usepackage{graphicx}
\usepackage{float}
\usepackage{pifont}
\usepackage{cleveref}
\usepackage{subcaption}
\usepackage[linesnumbered,ruled,lined,noend]{algorithm2e}
\usepackage{wrapfig}
\usepackage[]{minitoc}
\usepackage{thm-restate}
\usepackage{tabularx}
\usepackage{multirow}
\usepackage{array}
\usepackage{rotating}

\newcommand{\no}{\ding{55}}
\renewcommand{\vec}[1]{\bm{#1}}

\newcommand{\trans}{^{\!\top}}

\begin{document}

\title[Transferable Student Performance Modeling]{Transferable Student Performance Modeling \\ for Intelligent Tutoring Systems}

\author{Robin Schmucker}
\affiliation{%
  \institution{Carnegie Mellon University}
  \city{Pittsburgh}
  \state{Pennsylvania}
  \country{USA}}
\email{rschmuck@cs.cmu.edu}

\author{Tom M. Mitchell}
\affiliation{%
  \institution{Carnegie Mellon University}
  \city{Pittsburgh}
  \state{Pennsylvania}
  \country{USA}}
\email{tom.mitchell@cs.cmu.edu}

\renewcommand{\shortauthors}{Schmucker and Mitchell}

\begin{abstract}
Millions of learners worldwide are now using intelligent tutoring systems (ITSs). At their core, ITSs rely on machine learning algorithms to track each user's changing performance level over time to provide personalized instruction.
Crucially, student performance models are trained using interaction sequence data of \emph{previous} learners to analyse data generated by \emph{future} learners. This induces a \emph{cold-start} problem when a new course is introduced for which no training data is available.
Here, we consider transfer learning techniques as a way to provide accurate performance predictions for new courses by leveraging log data from existing courses. We study two settings: (i) In the \emph{naive transfer} setting, we propose \emph{course-agnostic} performance models that can be applied to any course. (ii) In the \emph{inductive transfer} setting, we tune pre-trained course-agnostic performance models to new courses using small-scale target course data (e.g., collected during a pilot study).
We evaluate the proposed techniques using student interaction sequence data from 5 different mathematics courses containing data from over 47,000 students in a real world large-scale ITS. The course-agnostic models that use additional features provided by human domain experts (e.g, difficulty ratings for questions in the new course) but no student interaction training data for the new course, achieve prediction accuracy on par with standard BKT and PFA models that use training data from thousands of students in the new course. In the inductive setting our transfer learning approach yields more accurate predictions than conventional performance models when only limited student interaction training data (<100 students) is available to both.
\end{abstract}

\begin{CCSXML}
<ccs2012>
<concept>
<concept_id>10010147.10010257</concept_id>
<concept_desc>Computing methodologies~Machine learning</concept_desc>
<concept_significance>500</concept_significance>
</concept>
<concept>
<concept_id>10010405.10010489.10010491</concept_id>
<concept_desc>Applied computing~Interactive learning environments</concept_desc>
<concept_significance>500</concept_significance>
</concept>
<concept>
<concept_id>10003456.10003457.10003527.10003540</concept_id>
<concept_desc>Social and professional topics~Student assessment</concept_desc>
<concept_significance>500</concept_significance>
</concept>
</ccs2012>
\end{CCSXML}

\ccsdesc[500]{Computing methodologies~Machine learning}
\ccsdesc[500]{Applied computing~Interactive learning environments}
\ccsdesc[500]{Social and professional topics~Student assessment}

\keywords{performance modeling, knowledge tracing, transfer learning}

\maketitle

\begin{figure}[!ht]
    \centering
    \includegraphics[width=0.475\textwidth]{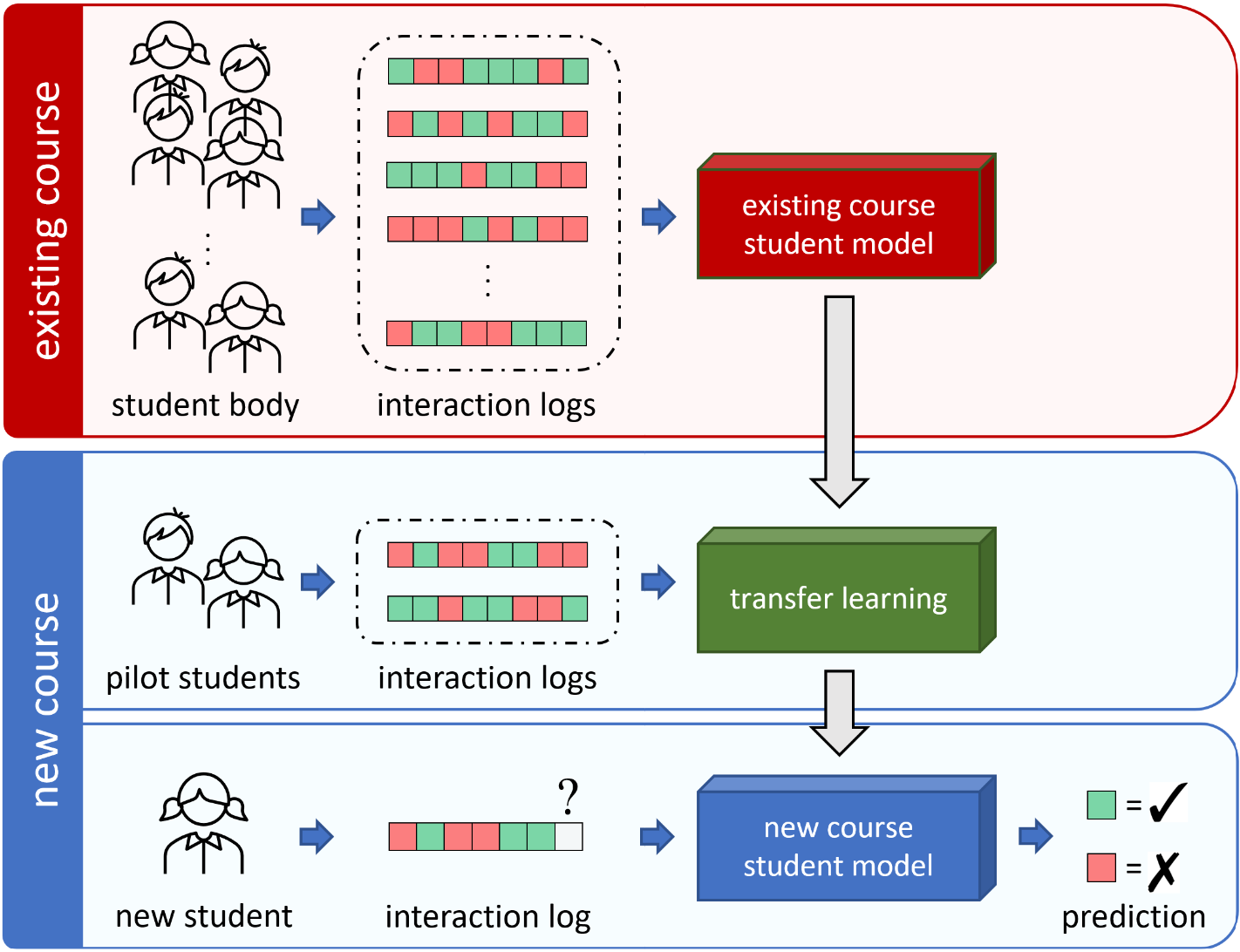}
    \caption{How can we use interaction log data from existing courses to train a student model for performance predictions in a new course for which only a little or no interaction log data is available?}
    \label{fig:concept}
\end{figure}

\section{Introduction}
\label{sec:introduction}

Intelligent tutoring systems (ITSs) are an educational technology which provides millions of learners worldwide with access to learning materials and personalized instruction. Even though ITS offerings come at a much lower cost, they can be nearly as effective as a personal human tutor in some cases~\cite{Vanlehn2011:Relative}. ITSs can be used to mitigate the academic achievement gap and help disadvantaged students~\cite{Huang2016:Intelligent}. At their core, ITSs rely on student performance models, that trace each student's changing ability level over time~\cite{Pelanek2017:Bayesian}, to adapt the curriculum to personal needs and to provide personalized feedback. Because of their crucial role, student performance models have received significant attention by the research community. 

The increasing demand for online tutoring systems induces a need for student performance modeling techniques that are flexible enough to support frequent releases of new courses, as well as changes to existing courses. The \emph{cold-start} problem, that arises when a new course is released for which no student log data is available for training, prevents us from applying conventional performance modeling approaches. In practice this means that the first batch of students does not enjoy the full benefits offered by the ITS. Future students then have the advantage that the log data generated by the early students can be used to train a student performance model that enables adaptive instruction and personalized feedback.

In this paper we consider transfer learning techniques as a way to improve the learning experience of early adopter students by mitigating the performance modeling cold-start problem for new courses. Transfer learning can be used to train accurate student performance models for a new (target) course by leveraging student log data collected from existing (source) courses (Figure~\ref{fig:concept}). We study two settings: (i) In the naive transfer setting where no data is yet available for the new target course, we learn \emph{course-agnostic} performance models that can be applied to any course, by using interaction sequence data from existing courses. Crucially, the features employed by our course-agnostic performance models avoid any dependencies on \emph{course-specific} attributes. (ii) In the inductive transfer setting where small-scale target course data is available, we tune pre-trained course-agnostic performance models to the target course by learning target course-specific question and knowledge component (KC) difficulty parameters. This inductive transfer setting mimics the case where the course designer can run a pilot with a small number of students before large-scale deployment.

We evaluate the proposed techniques using student log data collected from five different mathematics courses describing learning trajectories from over 47,000 students in a real world large-scale ITS. In both settings, the proposed transfer learning techniques mitigate the cold-start problem for all courses successfully. We hope that transfer learning techniques will become a standard tool for ITS course designers and improve the learning experience of early students. To summarize, the key contributions of this paper include:

\begin{itemize}

\item {\em Course-agnostic student performance modeling.} We present the first course-agnostic modeling techniques for predicting student performance on future questions in newly introduced courses where no previous students have yet taken this target course. Even though our agnostic models have no access to training data logs of students taking the new target course, they exhibit predictive performance comparable to conventional BKT and PFA student performance models -- found in many real world ITSs -- which were trained on data from thousands of students taking the new target course. Our course-agnostic models can enable effective personalized learning experiences even when introducing a new course for which no prior student log data is available. 

\item {\em Inductive transfer learning for effective tuning.} We use transfer learning techniques to efficiently adapt our pre-trained course-agnostic performance models to individual target courses by learning question- and KC-specific parameters. Our experiments show how our approach leads to more accurate performance predictions than conventional modeling techniques in settings in which only limited student log data from the target course is available (< 100 students).

\item {\em Guidance for practice.} By analyzing data from five different courses offered by a large-scale ITS this work provides various insights which can inform the design of future ITSs. Among others, our experiments show how manually assigned difficulty ratings and information about different learning contexts provided by domain experts during content creation can be used to boost the prediction accuracy of course-agnostic models. Further, going against common guidance, our analyses of various existing performance modeling techniques revealed that large logistic regression models can outperform classical lower dimensional models even in data starved settings (when training on <10 students). 
 
\end{itemize}

\section{Related Work}
\label{sec:related}

Here we start with a description of the transfer learning framework and discuss how it has been previously applied to multiple educational data mining (EDM) problems. We then provide an overview of existing student performance modeling approaches and discuss how they are affected by the cold-start problem.

\subsection{Transfer Learning}

Transfer learning techniques are a class of machine learning (ML) algorithms which aim to improve model performance in a \emph{target} domain (e.g., a new course) by leveraging data from a different but related \emph{source} domain (e.g., existing courses)~\cite{Zhuang2020:Comprehensive}. Transfer learning is particularly attractive when only limited target domain data is available, but source domain data is abundant. Via pre-training on source data, transfer learning can acquire a model for the target domain even when no target domain data is available. Transfer learning techniques enjoy great popularity in domains such as image classification~\cite{Yosinski2014:How} and machine translation~\cite{McCann2017:Learned}, but have also been applied to various educational data mining problems. 

In the context of learning management systems (LMS), transfer learning techniques that combine data from multiple different courses or from multiple offerings of the same course have been explored for predicting academic performance~\cite{Moreno2019:Generalizing, Lopez2020:Towards, Tsiakmaki2020:Transfer, Arroyo2021:Portability, Xing2021:Designing, Lopez2021:Improving, Ramaswami2022:Developing}. Going beyond predictions for individual courses there has been research on analysing data collected from multiple courses to predict the likelihood of passing future courses~\cite{Huynh2020:Integrating} and whether students will complete their degree program~\cite{Hunt2017:Transfer}. \citet{Gavsevic2016:Learning} studied the transferability of different models of academic success trained on LMS data collected by different courses and emphasize the importance that student models consider the course-specific context and its instructional conditions. 
In the setting of massive open online courses (MOOCs) transfer learning techniques that use data from previous offerings have been used to improve student dropout predictions in other offerings~\cite{Boyer2015:Transfer, Ding2019:Transfer} and the bias-variance trade-off of individual features has been studied~\cite{Kidzinsk2016:Generalizability}. 

Unlike all transfer approaches mentioned above, in this work we do not predict a single attribute related to the current course (i.e., pass/fail, student grade or dropout), but rather trace the changing likelihood with which students answer individual questions inside an ITS correctly over time based on their interaction history.

More related to the ITS setting considered in this paper, there have been multiple works that investigate how models that detect student gaming behaviour can be transferred between different courses and ITSs~\cite{Paquette2015:Cross, Paquette2018:System, Bahel2021:Transferring}. Using simulated students,~\citet{Spaulding2021:Towards} investigate a Gaussian Process-based approach for transferring cognitive models that describe learning word rhyming and spelling between different educational games for language learning. Multi-task learning techniques have been proposed to learn useful representations via pre-training on tasks related to response correctness and interaction time predictions for which large-scale training data is available~\cite{Choi2020:Assessment, Kim2021:Knowledge}. The learned representations are beneficial for the downstream TOIEC exam score predictions task which suffers from label scarcity.

\citet{Baker2019:Challenges} framed the problem of porting different types of student models (e.g., gaming detection models, performance models, \dots)  between different learning systems as an open challenge during a Keynote at the EDM2019 conference. In a recent update~\citet{Baker2021:Towards} survey related work and discuss potential directions for future research on sharing models across learning systems. While our work does not consider the transfer of student performance models across different ITSs, it focuses on the question of transferring performance models between different courses inside the \emph{same} ITS.

\subsection{Student Performance Modeling}

Student performance models estimate a student's ability to solve different questions based on sequential log data that describes their prior interactions with the system. The student proficiency estimates produced by such performance models are a key component, which allows the ITS to adapt to each student's personal needs as they go through the curriculum~\cite{Desmarais2012:Review}. In the literature there are three major categories of student performance modeling techniques: (i) Markov process based inference, (ii) logistic regression and (iii) deep learning based approaches. 
Markov process based techniques, such as Bayesian Knowledge Tracing (BKT)~\cite{Corbett1994:Knowledge} and BKT+~\cite{Khajah2016:Deep}, are well established and can for example be found in the Cognitive Tutor~\cite{Koedinger2006:Cognitive}. Most probabilistic approaches determine a student's proficiency level by performing inference in a two state Hidden Markov Model -- one state to represent mastery and one for non-mastery. 
Logistic regression models rely on a set of manually specified features which summarizes the student's interaction sequence. Given an input vector with feature values, the regression based performance model estimates the probability that the student is proficient in a certain question or KC. Some approaches in this class are IRT~\cite{Rasch1993:Probabilistic}, PFA~\cite{Pavlik2009:Performance}, DAS3H~\cite{Choffin2019:DAS3H}, Best-LR~\cite{Gervet2020:Deep} and its recent extension Best-LR+~\cite{Schmucker2021:Assessing}.
Deep learning based approaches take as input the same interaction sequence data, but unlike logistic regression techniques can learn suitable features on their own without requiring human feature engineering. Deep learning models benefit from large-scale training data and might in the future also include additional video and text data into their performance predictions. As of today, BKT and logistic regression models are still competitive with deep learning based approaches in multiple domains~\cite{Khajah2016:Deep, Gervet2020:Deep, Schmucker2021:Assessing}. Two comprehensive surveys on recent deep learning based performance models are provided by~\citet{Liu2021:Survey} and~\citet{Sarsa2021:Deep}.

Importantly, all of the above mentioned student performance modeling approaches rely on course-specific parameters (e.g., parameters that represent the difficulty of individual questions and KCs in the target course) which need to be learned from target course data. This makes these models inapplicable in our cold-start setting where a new course is first introduced and there is no training data available. In Subsection~\ref{subsec:naive_approach} we define a set of course-agnostic features which avoid any dependencies on course-specific attributes. This naive transfer approach allows us to learn course-agnostic student performance models using log data from existing courses that can be applied to any future courses.

Further, we want to mention recent works~\cite{Gervet2020:Deep, Zhang2021:Knowledge, Schmucker2021:Assessing} which investigated another cold-start problem related to student performance modeling. There, the question is \emph{how accurate} are the ability estimates of existing student performance models for new students for which we have only observed a few interactions. This is different from the cold-start problem studied in this paper -- it addresses the question of how to handle a new cold-start {\em student} in an existing course, whereas we address the question of how to handle a new cold-start {\em course}.  Related to the inductive transfer setting studied in this paper, is a short-paper by~\citet{Zhao2020:Cold} which addresses the cold-start problem by proposing an Attentive Neural Turing Machine architecture that requires less training data than an LSTM based approach. Unlike our study, they only experiment with small-scale student log data (<30 students, <1000 interactions) and their approach does not leverage data collected from existing courses for knowledge transfer.

\section{Setup}
\label{sec:setup}
Here, we start by providing a formal definition of student performance modeling as well as a description of the corresponding learning problem. We then introduce the multi-course dataset we use for our study, explain its student population and the properties which make the dataset suitable for our analysis.

\subsection{Formal Problem Statement}
\label{subsec:problem_statement}

\begin{table}[t]
    \small
    \centering
    \caption{Mathematical notation.}
    \begin{tabular}{c|l}
         \hline
         Notation & Description \\
         \hline
         $s$ & student index \\
         $t$ & time-step index \\
         $q_{s,t}$ & question answered by student $s$ at time $t$ \\
         $y_{s,t} \in \{0, 1\}$ & binary correctness of $s$ at time $t$ \\
         $c_{s,t}$ & additional context data for $s$ at time $t$ \\
         $x_{s,t} = (y_{s,t}, q_{s,t}, c_{s,t})$ & data for $s$ logged at time $t$ \\ 
         $\vec{x}_{s,1:t} = (x_{s,1}, \dots, x_{s,t})$ & all data for $s$ up to time $t$ \\
         $D = \{\vec{x}_{s_1,1:t_{1}}, \dots, \vec{x}_{s_n,1:t_n}\}$ & dataset containing logs from $n$ students \\
         $KC(q)$ & knowledge components targeted by $q$ \\ 
         $f_{\vec{w}}(\cdot)$ & performance model parameterized by $\vec{w}$ \\
         $L(\cdot, \cdot)$ & neg. log likelihood $L(a, b) = -\log(a \cdot b)$ \\
         \hline
    \end{tabular}
    \label{tab:notation}
\end{table}

We define a {\em student performance model} to be a function that takes as input the sequential log data from any student up to some point in the course, and produces as output a set of probabilities, where each probability is an estimate of how likely this student would answer correctly a specific question if asked that question at this point in the course.  Taken together, this collection of predicted probabilities for a particular collection of questions can be used as an estimate of the current knowledge state of the student. Accurate estimates allow the ITS to provide students with individualized feedback and enable adaptive learning sessions.

In this paper we consider the problem of {\em learning} such student performance models for a target course.  If many students have already completed the target course, we have a supervised learning problem in which we can use the log data from those students to train the student performance model.  If no students have yet taken the target course, we have a cold-start learning problem in which no such student data is available. Here we consider training the student performance model using student data from other courses.

Formally, we denote the sequence of student's $s$ past interaction with the system as  $\vec{x}_{s,1:t} = (x_{s,1}, \dots, x_{s,t})$. The tuple $x_{s,t} = (y_{s,t}, q_{s,t}, c_{s,t})$ represents the data collected for student $s$ at time-step $t$. Variable $q_{s,t}$ indicates the answered question, $y_{s,t} \in \{0, 1\}$ is binary response correctness and $c_{s,t}$ is an aggregation of additional information about question difficulty, learning context, read materials, watched videos and timestamp. Provided a student's interaction history $\vec{x}_{s,1:t}$ and a question identifier $q_{s, t + 1}$, a student performance model $f_{\vec{w}}$ estimates $p(y_{s, t + 1} = 1 \,|\, q_{s, t + 1}, \vec{x}_{s, 1:t}$) -- i.e. the probability that $s$ will respond correctly to $q_{s, t + 1}$ if it were asked next.

All performance models considered in this paper are parametric and defined by a weight vector $\vec{w} \in \mathbb{R}^d$. Using training data $D = \{\vec{x}_{s_1,1:t_{1}}, \dots, \vec{x}_{s_n,1:t_n}\}$ capturing interaction logs from previous students one can determine a vector $\vec{w}_D$ for predicting the performance of future students by solving the minimization problem
\begin{equation}
    \vec{w}_D = \arg \min_{\vec{w} \in \mathbb{R}^d} \sum_{s \in D} \sum_{t = 1}^{t_s} L(f_{\vec{w}}(q_{s,t}, \vec{x}_{s,1:t}), y_{s,t}).
\end{equation}
Here, $L(a, b) = -\log(a \cdot b)$ is the negative log likelihood function which penalizes model predictions $\hat{y}_{s,t} = f_{\vec{w}}(q_{s,t}, \vec{x}_{s,1:t})$ that deviate from the observed response correctness $y_{s,t}$. Gradient based optimization can used to solve this optimization problem. A summary of the mathematical notation is provided by Table~\ref{tab:notation}.

Following transfer learning nomenclature, $D_S = \{D_{S_1}, \dots, D_{S_k}\}$ is used to denote the \emph{source} data collected from existing courses $S_1, \dots, S_k$ and $D_T$ is the \emph{target} dataset from a new course $T$. When a new course is released a \emph{cold-start} problem arises because $D_T$ either contains no or only very little student interaction sequence data which prevents us from learning an accurate performance model $f_{\vec{w}_T}$ for the target course. In Section~\ref{sec:approach} we will propose transfer learning techniques that leverage log data from existing \emph{source} courses $S = \{S_1, \dots, S_k\}$ as a way to mitigate the cold-start problem for a new \emph{target} course $T$.

\subsection{Dataset}
\label{subsec:dataset}

For our analysis we rely on the \texttt{Squirrel Ai ElemMath2021} dataset~\cite{Schmucker2021:Assessing}. Squirrel Ai Learning (SQ-Ai) is an EdTech company located in China which offers personalized after-school tutoring services to K-12 students. Students can access SQ-Ai's ITS via mobile and web applications and select courses that target their desired subject and grade level. The ITS adapts to each individual student's needs by using machine learning techniques to estimate their ability level based on their interactions with the software. In addition to its online offerings, SQ-Ai also deploys its ITS in over 3000 physical tutoring centers in which students can study alongside their peers and work under the supervision of human teachers who provide additional guidance and support.

The \texttt{ElemMath2021} dataset captures log data from multiple mathematics courses for elementary school students and was collected over a 3 month period. Overall the dataset describes about 62,500,000 interactions from more than 125,000 students. Going beyond pure question-solving activities \texttt{ElemMath2021} also provides insights into how students interact with learning materials. During content creation domain experts assign each question a difficulty rating between 10 and 90. The domain experts also define a prerequisite graph which describes dependencies between individual KCs. \texttt{ElemMath2021} further records information about the learning context by assigning each learning activity to one of six categories of study modules (e.g. pre-test, effective learning, review, \dots).

Our study of the transferability of student performance models makes use of the fact that the \texttt{ElemMath2021} dataset is a combination of log data originating from different courses. Each student interaction is labeled with a course identifier which allows us to partition the logs into multiple course-specific datasets. For our analysis we selected the five courses with the most students, which we refer to as courses \texttt{C6}, \texttt{C7}, \texttt{C8}, \texttt{C9} and \texttt{C40}. Together, these five courses capture approximately 26,300,000 interactions from over 47,000 students. Table~\ref{tab:data_courses} shows statistics for the individual courses. On average, students answer about 200 questions in a single course and the correctness rate varies from course to course between 62.4\% and 71.3\%. Each \texttt{ElemMath2021} student only participates in a single course which implies \emph{disjoint} student populations across courses. In terms of covered KCs and used questions the courses are also \emph{disjoint} with the exception of \texttt{C9} and \texttt{C40} which have an overlap of less than 5\%. These properties allow us to measure the transferability of student performance models to different courses involving disjoint students and disjoint questions and knowledge components.

\begin{table}[t]
    \centering
    \caption{Five largest \texttt{ElemMath2021} courses by students.}
    \begin{tabular}{l|c|c|c|c|c}
    \hline
         course & \texttt{C6} & \texttt{C7} & \texttt{C8} & \texttt{C9} & \texttt{C40} \\
         \hline
         \# of students & 11,864 & 9,423 & 10,296 & 8,531 & 7,487 \\
         \# of questions & 2,483 & 2,226 & 2,438 & 2,407 & 1,307 \\
         \# of KCs & 164 & 145 & 159 & 157 & 87 \\
         \# of logs & 8,212k & 5,576k & 5,112k & 3,767k & 3,614k \\
         \# of responses & 3,262k & 1,934k & 2,142k & 1,407k & 1,228k \\
         \hline
         avg. resp. & 275 & 227 & 187 & 165 & 164 \\
         avg. correct & 71.30\% & 69.62\% & 69.47\% & 68.68\% & 62.39\% \\
         \hline
    \end{tabular}
    \label{tab:data_courses}
\end{table}

\section{Approach}
\label{sec:approach}

We investigate two transfer learning approaches to mitigate the student performance modeling problem for new courses by leveraging student log data from existing courses. First, in the \emph{naive transfer} setting we identify a set of course-agnostic features that can be used to train general performance models that can predict student ability for any course. Second, in the \emph{inductive transfer} setting we show how one can tune a pre-trained course-agnostic performance model to a specific target course using only very limited student log data. The inductive transfer setting captures the case where the course designer can run a pilot study with a small number of students before large-scale deployment.

\subsection{Naive Transfer}
\label{subsec:naive_approach}

The naive transfer setting is concerned with leveraging student log data $D_S$ from existing source courses $S = \{S_1, \dots, S_k\}$ to learn a student performance model that is general enough to be applied to any future target course $T$. Crucially, such a \emph{course-agnostic} performance modeling approach cannot rely on any parameters that describe course-specific elements. Because existing student performance modeling techniques rely on parameters that capture properties of individual questions and KCs, they require access to target course data $D_T$ for training and are thus not applicable when training data is not available.

As a first step in the design of course-agnostic performance models we identify a set of general features that do not induce the need for course-specific training data. For this we study existing logistic regression based performance modeling techniques. At the core of each regression model lies a feature function $\Phi = (\phi_1, \dots, \phi_d)$ which outputs a real-valued vector that describes student $s$'s prior interaction history $\vec{x}_{s,1:t}$ as well as information about the next question $q_{s,t+1}$. Each logistic regression approach is characterized by its own feature function which computes a set of distinct features to summarize the interaction history. The trained logistic regression model then uses this feature vector as input to estimate the probability that $s$ will respond correctly to question $q_{s,t+1}$ if it were asked next. The corresponding performance prediction is defined as 
\begin{equation}
  p(y_{s, t+1} = 1 \,|\, q_{s, t+1}, \vec{x}_{s, 1:t}) = \sigma \left( \vec{w}\trans \Phi(q_{s, t+1}, \vec{x}_{s,1:t}) \right).
\end{equation}
 Here $\vec{w} \in \mathbb{R}^d$ is the learned weight vector that defines the model and $\sigma(x) = 1 / (1 + e^{-x}) \in [0, 1]$ is the sigmoid function whose output can be interpreted as the probability of correct response. A suitable set of regression weights can be learned using training data from previous students as described in Subsection~\ref{subsec:problem_statement}.

 Because conventional student performance modeling techniques use feature functions that produce course-specific elements they do not generalize to new courses. As an example consider the Best-LR model by~\citet{Gervet2020:Deep}. It features a student ability parameter $\alpha_s$ and difficulty parameters $\delta_q$ and $\beta_k$ for each individual question $q$ and knowledge component $k$.
Further, Best-LR uses count features to track the number of prior correct ($c_s$) and incorrect ($f_s$) responses of student $s$ overall and for each individual KC $k$ (i.e., $c_{s, k}$ and $f_{s, k}$). Defining $\phi(x) = \log(1 + x)$, the Best-LR prediction is
\begin{equation}
\label{eq:bestlr}
\begin{split}
  p_{\text{Best-LR}}(y_{s, t+1} =  1 \,|\, q_{s, t+1}, \vec{x}_{s, 1:t}) &= 
  \sigma ( \alpha_s - \delta_{q_{s, t+1}} + \tau_c \phi(c_s)\, + \\ \tau_f \phi(f_s)  + 
  \sum_{k \in KC(q_{s, t+1})} \beta_k + &\gamma_k \phi(c_{s, k}) + \rho_k \phi(f_{s, k})).
\end{split}
\end{equation}
One can interpret the Best-LR feature function as a tuple $\Phi = (\Phi_A, \Phi_T)$ where $\Phi_A$ is course-agnostic (i.e., student ability, and total count features) and $\Phi_T$ is target course-specific (i.e., question and skill difficulty and count features). Because -- to the best of our knowledge -- this is the first work that investigates the problem of course-agnostic student performance modeling we introduce simple but reasonable baselines by taking conventional performance modeling approaches and reducing them to their respective course-agnostic feature set $\Phi_A$. Note that the avoidance of course-specific features reduces model expressiveness and predictive performance considerably (discussed in Subsection~\ref{subsec:inductive_approach}).

Looking again at the Best-LR example we derive a course-agnostic student performance model called A-Best-LR. A-Best-LR only relies on student ability and overall count features $c_s$ and $f_s$. In addition, it employs two parameters $\gamma$ and $\rho$ to consider the number of prior correct ($c_{s, k}$) and incorrect responses ($f_{s, k}$) related to the current KC $k$ -- the same $\gamma$ and $\rho$ parameters are used for \emph{all} KCs. The A-Best-LR prediction is defined as
\begin{align}
\label{eq:agnostic_bestlr}
\begin{split}
  p_{\text{A-Best-LR}}&(y_{s, t+1} =  1 \,|\, q_{s, t+1}, \vec{x}_{s, 1:t}) =  \\
  &\sigma (\alpha_s + \tau_c \phi(c_s) + \tau_f \phi(f_s) + \gamma \phi(c_{s, k}) + \rho \phi(f_{s, k})).
\end{split}
\end{align}

By avoiding course-specific features the A-Best-LR model can be trained on source data $D_S$ collected from existing courses and then be used for any new course $T$. Giving a similar treatment to other popular performance modeling techniques we define:
\begin{itemize}
    \item A-BKT: We train a single BKT~\cite{Corbett1994:Knowledge} parameter set to model performance for all students and KCs. During deployment we estimate student performance by using the learned parameters to initialize BKT models for each individual KC. 
    \item A-IRT: We train an IRT (Rasch) model~\cite{Rasch1993:Probabilistic} that uses the same difficulty parameter for all questions. We then use this difficulty parameter to trace student ability over time for each KC and derive performance predictions.
    \item A-PFA: We train a simplified 3-parameter PFA model~\cite{Pavlik2009:Performance} that uses the same difficulty, correctness and incorrectness count parameters for all KCs.
    \item A-DAS3H: We train a simplified DAS3H model~\cite{Choffin2019:DAS3H} that uses a shared difficulty parameter for all questions and KCs and that also uses a single a set of time-window based correctness and incorrectness count parameters for all KCs.  
    \item A-Best-LR+: We train a simplified Best-LR+ model~\cite{Schmucker2021:Assessing} that augments the A-Best-LR feature set (EQ~\ref{eq:agnostic_bestlr}) with response pattern and smoothed average correctness features. In addition, the model learns a single set of DAS3H~\cite{Choffin2019:DAS3H} time-window and R-PFA~\cite{Galyardt2015:Move} and PPE~\cite{Walsh2018:Mechanisms} count parameters used for all KCs.
\end{itemize}

Conventional student modeling approaches -- including all of the above -- base their performance predictions mainly on two types of features: (i) Question and KC one-hot encodings that allow the model to learn distinct difficulty parameters for each question and KC; (ii) Count features that summarize a student's prior interactions with the ITS. Recently, it has been shown how alternative types of log data collected by modern tutoring systems can be incorporated into logistic regression models to improve performance predictions~\cite{Schmucker2021:Assessing}. The use of such alternative types of features is particularly interesting in the naive transfer setting because most conventional features are course-specific and thus cannot be used for new courses. The \texttt{ElemMath2021} dataset (see Subsection~\ref{subsec:dataset}) offers various types of student interaction data. In our experiments (Section~\ref{sec:experiments}) we consider information related to student video and reading material consumption, current learning context, question difficulty ratings assigned by human domain experts during content creation, KC prerequisite structure as well as response- and lag-time features introduced by SAINT+~\cite{Shin2021:Saint+}. By augmenting the A-Best-LR+ feature set with some these alternative features we are able to use data from previous courses $D_S$ to train course-agnostic performance models whose prediction accuracy is on par with standard BKT and PFA models trained on target course data $D_T$.

\subsection{Inductive Transfer}
\label{subsec:inductive_approach}

Most conventional student performance modeling approaches rely on parameters that capture question- and KC-specific attributes. By training and testing on target course data $D_T$ using a 5-fold cross validation, Table~\ref{tab:advantage_item_skill} compares the performance of course-agnostic performance models with models that use the same course-agnostic feature set, but which are allowed to learn additional course-specific parameters to capture question- and KC-difficulty. We observe that the inclusion of question- and KC-specific parameters leads to large improvements in prediction accuracy and closes the gap to conventional student performance modeling techniques (Table~\ref{tab:traditional}).

\begin{table}[t]
    \centering
    \caption{When training and testing on data from the \emph{same} course, adding course-specific features to the course-agnostic A-AugLR feature set results in much higher accuracy. The A-AugLR row shows average model performance when using only course-agnostic features. The next three rows show the performance of A-AugLR models with added KC- (+KC) and question-specific (+quest) parameters.}
    \begin{tabular}{l|c|c}
    \hline
           & ACC & AUC \\
         \hline
         Always correct &  68.29 & 50.00 \\
         \hline
         A-AugLR &  72.02 & 69.48 \\
         A-AugLR+KC & 74.00 & 74.99 \\
         A-AugLR+quest. & 76.34 & 79.39 \\
         A-AugLR+KC+quest. & \textbf{76.37} & \textbf{79.39} \\
         \hline
    \end{tabular}
    \label{tab:advantage_item_skill}
\end{table}

Motivated by this, we propose an inductive transfer learning approach that uses small-scale target course data $D_T$ to tune a pre-trained course-agnostic performance model to a new course $T$ by learning additional question- and KC-specific parameters. Formally, the agnostic and target model are parameterized by weight vectors $\vec{w}_S \in \mathbb{R}^{|\Phi_S|}$ and $\vec{w}_T \in \mathbb{R}^{|\Phi_S| + |\Phi_T|}$ respectively. Similar to ~\citet{Orabona2009:Model} we use $L_2$ regularization to subject the target weight vector $\vec{w}_T$ to a Gaussian prior $\mathcal{N}((\vec{w}_S, \vec{0})\trans, \, \vec{1})$. We control the degree of regularization using a penalty parameter $\lambda \in \mathbb{R}_{\geq 0}$. The corresponding regularized maximum likelihood objective is
\begin{equation}
    \vec{w}_T = \arg \min_{\vec{w} \in \mathbb{R}^d} \frac{\lambda}{2} \|\vec{w} - 
    \begin{pmatrix} \vec{w}_S \\ \vec{0} \end{pmatrix}
     \|_2^2 + \sum_{s \in D_T} \sum_{t = 1}^{t_s} L(f_{\vec{w}}(q_{s,t}, \vec{x}_{s,1:t}), y_{s,t}).
\end{equation}
By using a prior for $w_T$ that is based on the previously learned $w_S$, we can mitigate overfitting and can learn a suitable target model using only very limited training data $D_T$. With increasing amounts of recorded learning histories in $D_T$ the objective focuses increasingly on model fit. For our experiments we determine the penalty parameter value by evaluating $\lambda \in \{0.01, 0.05, 0.1, 0.5, 1, 5, 10, 100 \}$ using the first split of a 5-fold cross validation on the \texttt{C6} training data. We found $\lambda = 5$ to be most effective for different amounts of tuning data and use it for all our experiments.

\section{Experiments}
\label{sec:experiments}

We evaluate the proposed transfer learning techniques using student interaction sequence data from five different mathematics courses taken from the \texttt{ElemMath2021} dataset (Subsection~\ref{subsec:dataset}). In the naive transfer setting we first evaluate the utility of different features and then select a set of features that can be used to train accurate course-agnostic student performance models. In the inductive transfer setting we show our approach yields more accurate performance predictions than conventional modeling approaches when only small-scale student log data is available to both.

\subsection{Evaluation Methodology}
\label{subsec:evaluation_methodology}

\begin{table*}[!ht]
\small
\centering
\caption{Reference model performance. Average ACC and AUC metrics achieved by conventional \emph{course-specific} student performance models that were trained and tested on data from the \emph{same} course. The largest observed ACC and AUC variances over the five-fold test data are both 0.01\%.}
\begin{tabular}{l|cc|cc|cc|cc|cc|cc}
\hline
& \multicolumn{2}{c|}{\texttt{C6}} & \multicolumn{2}{c|}{\texttt{C7}} & \multicolumn{2}{c|}{\texttt{C8}} & \multicolumn{2}{c|}{\texttt{C9}} & \multicolumn{2}{c|}{\texttt{C40}} & \multicolumn{2}{c}{Average} \\
model \,\textbackslash\, in \%    & ACC & AUC & ACC & AUC & ACC & AUC & ACC & AUC & ACC & AUC & ACC & AUC\\
\hline
Always correct & 71.30 & 50.00 & 69.62 & 50.00 & 69.47 & 50.00 & 68.68 & 50.00 & 62.38 & 50.00 & 68.29 & 50.00 \\
\hline
BKT & 74.89 & 73.39 & 71.66 & 69.35 & 72.24 & 70.43 & 72.01 & 70.09 & 68.09 & 71.00 & 71.78 & 70.85 \\
PFA & 74.66 & 73.02 & 71.52 & 69.19 & 72.13 & 70.21 & 71.87 & 69.94 & 67.85 & 70.87 & 71.61 & 70.65 \\
IRT & 75.52 & 75.66 & 73.05 & 73.22 & 73.28 & 73.21 & 72.40 & 72.36 & 68.66 & 72.05 & 72.58 & 73.30 \\
\hline
DAS3H        & 77.31 & 78.15 & 74.59 & 76.06 & 75.05 & 76.18 & 74.09 & 75.38 & 70.87 & 75.20 & 74.38 & 76.19 \\
Best-LR      & 78.42 & 80.30 & 75.95 & 78.44 & 76.58 & 78.97 & 76.33 & 79.08 & 73.10 & 78.07 & 76.08 & 78.97 \\
Best-LR+     & \textbf{78.75} & \textbf{80.85} & \textbf{76.18} & \textbf{78.83} & \textbf{76.90} & \textbf{79.39} & \textbf{76.69} & \textbf{79.58} & \textbf{73.62} & \textbf{78.81} & \textbf{76.43} & \textbf{79.49} \\
\hline
A-AugLR & 74.40 & 69.90 & 72.05 & 67.80 & 72.45 & 68.49 & 72.83 & 70.80 & 68.38 & 70.42 & 72.02 & 69.48 \\
A-AugLR+KC+quest & 78.66 & 80.74 & 76.08 & 78.69 & 76.86 & 79.25 & 76.68 & 79.54 & 73.57 & 78.72 & 76.37 & 79.39 \\
\hline
\end{tabular}
\label{tab:traditional}
\end{table*}

As is common in prior work~\cite{Piech2015:Deep, Choffin2019:DAS3H, Gervet2020:Deep, Schmucker2021:Assessing} we filter out students with less than ten answered questions. In the naive transfer setting, we use each course once to simulate a new target course $T \in \{\texttt{C6}, \texttt{C7}, \texttt{C8}, \texttt{C9}, \texttt{C40}\}$. For each target $T$ we train one course-agnostic performance model using data from the other four courses and then evaluate prediction on unseen target dataset $D_T$. For the inductive transfer experiments we perform a 5-fold cross-validation on the student level where in each fold 80\% of students are used as training set $D_{T,\text{train}}$ and the remaining 20\% are used as test set $D_{T,\text{test}}$. To simulate small-scale training data we sample a limited number of students (5, 10, \dots) from training set $D_{T,\text{train}}$. Because the Squirrel Ai tutoring system tends to introduce course topics in the same sequential order we only select students that reached the last topic -- selected students might have skipped or revisited individual topics. This approach mimics the case where the course designer can collect interaction log data from a small number of students during a pilot study before large-scale deployment. We report model performance using accuracy (ACC) and area under curve (AUC) metrics.

Our code builds on the public GitHub repository by~\citet{Schmucker2021:Assessing} which implements various features and state-of-the-art logistic regression models. We follow their Best-LR+ model and set smoothing parameter $\eta = 5$, response pattern sequence length $n = 10$, R-PFA ghost attempts $g = 3$, R-PFA decay rates $d_F = 0.8$ and $d_R = 0.8$, and PPE parameters $c = 1$, $x = 0.6$, $b = 0.01$ and $m = 0.028$. We implemented the BKT experiments using pyBKT~\cite{Badrinath2021:Pybkt}. For our naive and inductive transfer experiments we rely on PyTorch~\cite{Paszke2019:Pytorch} and train each regression model for 200 epochs using the Adam optimizer~\cite{Kingma2014:Adam} with learning-rate $\alpha = 0.001$. As a reference, Table~\ref{tab:traditional} provides average performance metrics of conventional student performance modeling approaches that were trained and tested on the \emph{same} course using a 5-fold cross-validation.

\subsection{Naive Transfer}
\label{subsec:naive_experiments}

\begin{table*}[t]
\small
\centering
\caption{Naive transfer feature evaluation. We used each of the five courses to simulate a new target course and trained \emph{course-agnostic} student performance models using the A-BestLR+ feature set augmented with one additional feature on data from the other four courses. We then evaluated AUC and ACC performance on the new target course. The marker \no\, indicates which additional features produced the greatest improvements, and were therefore selected for the A-AugLR model.}
\begin{tabular}{l|cc|cc|cc|cc|cc|ccc}
\hline
 & \multicolumn{2}{c|}{\texttt{C6}} & \multicolumn{2}{c|}{\texttt{C7}} & \multicolumn{2}{c|}{\texttt{C8}} & \multicolumn{2}{c|}{\texttt{C9}} & \multicolumn{2}{c|}{\texttt{C40}} & \multicolumn{2}{c}{Average} & \\
feature \,\textbackslash\, in \%    & ACC & AUC & ACC & AUC & ACC & AUC & ACC & AUC & ACC & AUC & ACC & AUC & \\
\hline
A-BestLR+ (baseline) & 73.86 & 67.33 & 71.62 & 65.41 & 71.92 & 65.79 & 72.36 & 68.92 & 67.59 & 68.31 & 71.47 & 67.15 & \\
\hline
current lag time  & 73.91 & 67.48 & 71.55 & 65.42 & 72.00 & 66.00 & 72.40 & 69.02 & 67.65 & 68.38 & 71.50 & 67.26 & \no \\
prior response time  & 73.94 & 67.61 & 71.57 & 65.39 & 72.02 & 65.98 & 72.39 & 69.12 & 67.46 & 68.41 & 71.48 & 67.30 & \no \\
\hline
learning context one-hot  & 73.83 & 67.24 & 71.65 & 65.56 & 71.94 & 65.95 & 72.38 & 69.02 & 67.65 & 68.70 & 71.49 & 67.29 & \no \\
learning context counts  & 73.87 & 67.38 & 71.53 & 65.33 & 72.04 & 65.79 & 72.41 & 69.09 & 67.71 & 68.54 & 71.51 & 67.23 & \no \\
difficulty one-hot & 74.09 & 68.63 & 71.88 & 66.80 & 72.22 & 67.21 & 72.54 & 69.71 & 67.82 & 68.84 & 71.71 & 68.24 & \no \\
difficulty counts  & 73.84 & 67.34 & 71.60 & 65.56 & 71.93 & 66.00 & 72.33 & 69.00 & 67.59 & 68.52 & 71.46 & 67.28 & \no \\
\hline
prereq counts  & 73.88 & 67.27 & 71.58 & 65.44 & 71.91 & 65.83 & 72.31 & 68.92 & 67.55 & 68.28 & 71.45 & 67.15 & \\
postreq counts  & 73.61 & 66.48 & 71.68 & 65.03 & 71.98 & 65.96 & 72.38 & 69.22 & 67.39 & 68.15 & 71.41 & 66.97 & \\
\hline
videos watched counts  & 73.84 & 67.32 & 71.59 & 65.41 & 71.95 & 65.75 & 72.30 & 68.91 & 67.52 & 68.20 & 71.44 & 67.12 & \\
reading counts & 73.83 & 67.41 & 71.60 & 65.48 & 71.96 & 65.82 & 72.37 & 68.93 & 67.53 & 68.19 & 71.46 & 67.17 & \\
\hline
\end{tabular}
\label{tab:feature_eval_bestlrp_holdout}
\end{table*}

\subsubsection{Feature Evaluation} We evaluate the benefits of the alternative features discussed in Subsection~\ref{subsec:naive_approach} for course-agnostic student performance modeling. For each feature, we train an augmented A-Best-LR+ model using the A-Best-LR+ feature set plus one additional feature. We rely on A-Best-LR+ because it builds on a combination of features derived from various other performance models and yields the most accurate performance predictions among all considered course-agnostic baseline models (Table~\ref{tab:naive_transfer_holdout}). 

Table~\ref{tab:feature_eval_bestlrp_holdout} shows the ACC and AUC scores when using different features. The one-hot features that encode the difficulty ratings assigned by domain experts during content creation yield the largest ACC and AUC improvements over the A-Best-LR+ baseline for all courses -- on average $0.24\%$ for ACC and $1.07\%$ for AUC. The one-hot features that encode the learning context a question is placed in, improve the average AUC score by $0.14\%$. The count features that track the number of prior correct and incorrect responses to questions of a certain difficulty or learning context, lead to smaller improvements compared to their one-hot counterparts. The lag time and response time features introduced by SAINT+\cite{Shin2021:Saint+} improve AUC scores on average by $0.15\%$ and $0.11\%$. The post- and pre-requisite features that were extracted from the KC dependency graph did not benefit the course-agnostic performance predictions. It is not clear how to learn prerequisite graph parameters that generalize to new courses. Similarly, the count features that summarize the students' video and reading material usage did not improve the performance predictions. One limitation of these two features is that they do not capture the relationship between the content covered by different learning materials and individual questions.

\begin{table*}[t]
\small
\centering
\caption{Naive transfer performance. We used each of the five courses to simulate a new target course and trained \emph{course-agnostic} performance models using student interaction data from the other four courses. We then evaluated AUC and ACC performance on the new target course. We highlight the fact that these models are stripped from all course-specific parameters (e.g., question- and KC- difficulty) and can be used to analyze student interaction data from any course.}
\begin{tabular}{l|cc|cc|cc|cc|cc|cc}
\hline
& \multicolumn{2}{c|}{\texttt{C6}} & \multicolumn{2}{c|}{\texttt{C7}} & \multicolumn{2}{c|}{\texttt{C8}} & \multicolumn{2}{c|}{\texttt{C9}} & \multicolumn{2}{c|}{\texttt{C40}} & \multicolumn{2}{c}{Average} \\
model \,\textbackslash\, in \%    & ACC & AUC & ACC & AUC & ACC & AUC & ACC & AUC & ACC & AUC & ACC & AUC\\
\hline
Always correct & 71.30 & 50.00 & 69.62 & 50.00 & 69.47 & 50.00 & 68.68 & 50.00 & 62.38 & 50.00 & 68.29 & 50.00  \\
\hline
A-BKT      & 73.25 & 63.29 & 70.58 & 60.87 & 71.04 & 60.85 & 70.93 & 62.30 & 65.56 & 63.87 & 70.27 & 62.57 \\
A-PFA & 73.27 & 63.54 & 70.75 & 60.59 & 71.13 & 61.23 & 70.92 & 62.29 & 65.55 & 63.13 & 70.32 & 62.16 \\
A-IRT & 59.50 & 61.82 & 58.36 & 59.55 & 57.50 & 59.45 & 58.04 & 60.96 & 58.33 & 62.72 & 58.35 & 60.90 \\
\hline
A-DAS3H & 73.29 & 63.70 & 70.81 & 60.84 & 71.15 & 61.31 & 70.98 & 62.41 & 65.59 & 63.54 & 70.36 & 62.36 \\
A-Best-LR & 73.55 & 66.38 & 71.35 & 64.17 & 71.74 & 65.01 & 71.99 & 67.87 & 67.13 & 66.86 & 71.15 & 66.06 \\
A-Best-LR+ & 73.86 & 67.33 & 71.62 & 65.41 & 71.92 & 65.79 & 72.36 & 68.92 & 67.59 & 68.31 & 71.47 & 67.15 \\
\hline
A-AugLR & \textbf{74.28} & \textbf{69.11} & \textbf{71.80} & \textbf{67.21} & \textbf{72.35} & \textbf{68.19} & \textbf{72.76} & \textbf{70.52} & \textbf{68.05} & \textbf{69.52} & \textbf{71.85} & \textbf{68.91} \\
\hline
\end{tabular}
\label{tab:naive_transfer_holdout}
\end{table*}

\subsubsection{Agnostic AugmentedLR} We have identified a set of features that individually improve the performance predictions made by the A-Best-LR+ model (highlighted in Table~\ref{tab:feature_eval_bestlrp_holdout}). Inspired by the recent AugmentedLR paper~\cite{Schmucker2021:Assessing}, we propose a course-agnostic A-AugLR model by augmenting the A-Best-LR+ feature set with these highlighted features: lag and response time, learning context and question difficulty features. Performance metrics for A-AugLR and the naive transfer baselines defined in Subsection~\ref{subsec:naive_approach} are provided by Table~\ref{tab:naive_transfer_holdout}. We observe that the course-agnostic models derived from BKT, PFA, IRT and DAS3H struggle in the naive transfer setting and yield low AUC scores. Our A-IRT implementation that only uses a single question difficulty parameter suffers from fluctuating student ability estimates and low prediction accuracy. The A-Best-LR feature set contains count features that describe the student's overall number of prior correct and incorrect responses which provides an advantage over the A-DAS3H model. A-Best-LR+ uses various additional features to capture aspects of long- and short-term student performance over time. On average, its predictions yield $0.37\%$ higher ACC and $1.09\%$ higher AUC scores compared to the A-Best-LR model it builds on.

The A-AugLR model -- which augments A-Best-LR+ with multiple features of which each has been found to improve predictions on its own (those highlighted in Table~\ref{tab:feature_eval_bestlrp_holdout}) -- yields the best performance predictions in the naive transfer setting. Compared to A-Best-LR+, the A-AugLR models are on average $0.38\%$ more accurate and their AUC scores are $1.76\%$ higher. This emphasizes how additional information provided by human domain experts during content creation can enable accurate performance predictions in the naive transfer setting. Importantly, even though our course-agnostic A-AugLR models were fitted using data from different courses their prediction accuracy is on par with course-specific BKT and PFA models which were trained on target course data (compare the  A-AugLR row in Table~\ref{tab:naive_transfer_holdout} to the BKT and PFA rows from Table~\ref{tab:traditional}). This shows how A-AugLR can mitigate the cold-start problem for new courses even for the very first student.

\begin{table}[t]
    \small
    \centering
    \caption{A-AugLR: ACC metrics achieved when training and testing on different source-target course pairs.}
    \begin{tabular}{l|c|c|c|c|c}
    \hline
         train / test & \texttt{C6} & \texttt{C7} & \texttt{C8} & \texttt{C9} & \texttt{C40} \\
         \hline
         \texttt{C6} &  \textbf{74.43} & 71.82 & 72.25 & 72.53 & 67.83 \\
         \texttt{C7} &  73.98 & \textbf{72.09} & 72.32 & 72.61 & 67.96 \\
         \texttt{C8} &  74.25 & 71.92 & \textbf{72.55} & 72.73 & 67.86 \\
         \texttt{C9} &  74.23 & 71.88 & 72.43 & \textbf{72.93} & 68.03 \\
         \texttt{C40} & 73.87 & 71.77 & 71.97 & 72.41 & \textbf{68.54} \\
         \hline
    \end{tabular}
    \label{tab:transerability_acc}
\end{table}

\begin{table}[t]
    \small
    \centering
    \caption{A-AugLR: AUC metrics achieved when training and testing on different source-target course pairs.}
    \begin{tabular}{l|c|c|c|c|c}
    \hline
         train / test & \texttt{C6} & \texttt{C7} & \texttt{C8} & \texttt{C9} & \texttt{C40} \\
         \hline
         \texttt{C6} &  \textbf{70.00} & 66.91 & 67.70 & 69.88 & 69.26 \\
         \texttt{C7} &  68.75 & \textbf{67.95} & 67.99 & 70.16 & 69.45 \\
         \texttt{C8} &  68.90 & 67.05 & \textbf{68.68} & 70.49 & 68.80 \\
         \texttt{C9} &  68.76 & 66.73 & 67.90 & \textbf{71.07} & 69.28 \\
         \texttt{C40} & 67.81 & 66.54 & 66.69 & 69.60 & \textbf{70.68} \\
         \hline
    \end{tabular}
    \label{tab:transerability_auc}
\end{table}

\subsubsection{Single course transfer} So far we learned course-agnostic student performance models for each individual target course $T \in \{\texttt{C6}, \texttt{C7}, \texttt{C8}, \texttt{C9}, \texttt{C40}\}$ by training on combined log data from the other four courses. This raises the question if instead of using data from multiple courses, one should try to identify and train on the single course that is most similar to the target domain. Tables~\ref{tab:transerability_acc} and~\ref{tab:transerability_auc} try to answer this question by training and testing course-agnostic A-AugLR models on different pairs of courses. 

Unsurprisingly, the A-AugLR models that are trained and tested on the same course exhibit the highest ACC and AUC scores (diagonal entries). Compared to these optimal A-AugLR models -- which require target course data for training -- the accuracy gaps between the most and least compatible course pairs vary between $0.12\%$ (\texttt{C8}/\texttt{C9}) and $0.71\%$ (\texttt{C6}/\texttt{C40}). The AUC score gaps are larger and vary between $0.58\%$ (\texttt{C8}/\texttt{C9}) and $2.19\%$ (\texttt{C6}/\texttt{C40}). The \texttt{C6}/\texttt{C40} pair is least compatible for naive transfer. One possible reason for this is the fact that \texttt{C6} and \texttt{C40} exhibit the highest ($71.3\%$) and lowest ($62.4\%$) average correctness rates among all courses. The \texttt{C8}/\texttt{C9} pair is most compatible for naive transfer. Both exhibit similar average correctness rates ($69.5\%$/$68.7\%$), number of KCs (159/157) and students answer a similar number of questions on average (187/165).

Next, we compare the predictive performance of the A-AugLR models that were trained and tested on the target course (diagonals of Tables~\ref{tab:transerability_acc} and~\ref{tab:transerability_auc}) with the A-AugLR models that were trained using data from the other four courses (Table~\ref{tab:naive_transfer_holdout}). Here the average ACC and AUC performance gaps are $0.26\%$ and $0.77\%$ respectively. Overall, we conclude that for the courses considered in this paper the exact course pairing tends to make make little difference in naive transfer performance. There is no clear criterion for selecting the single best course for transfer, also because we do not know the target course correctness rate before deployment. Combining the data from all non-target courses for training proved itself to be a suitable strategy. We emphasize that all considered courses cover mathematics topics for elementary school students. One might observe larger differences in transfer performance when analysing a more diverse set of courses.

\subsection{Inductive Transfer}
\label{subsec:inductive_experiments}

\begin{figure*}[!ht]
    \centering
    \includegraphics[width=0.435\textwidth]{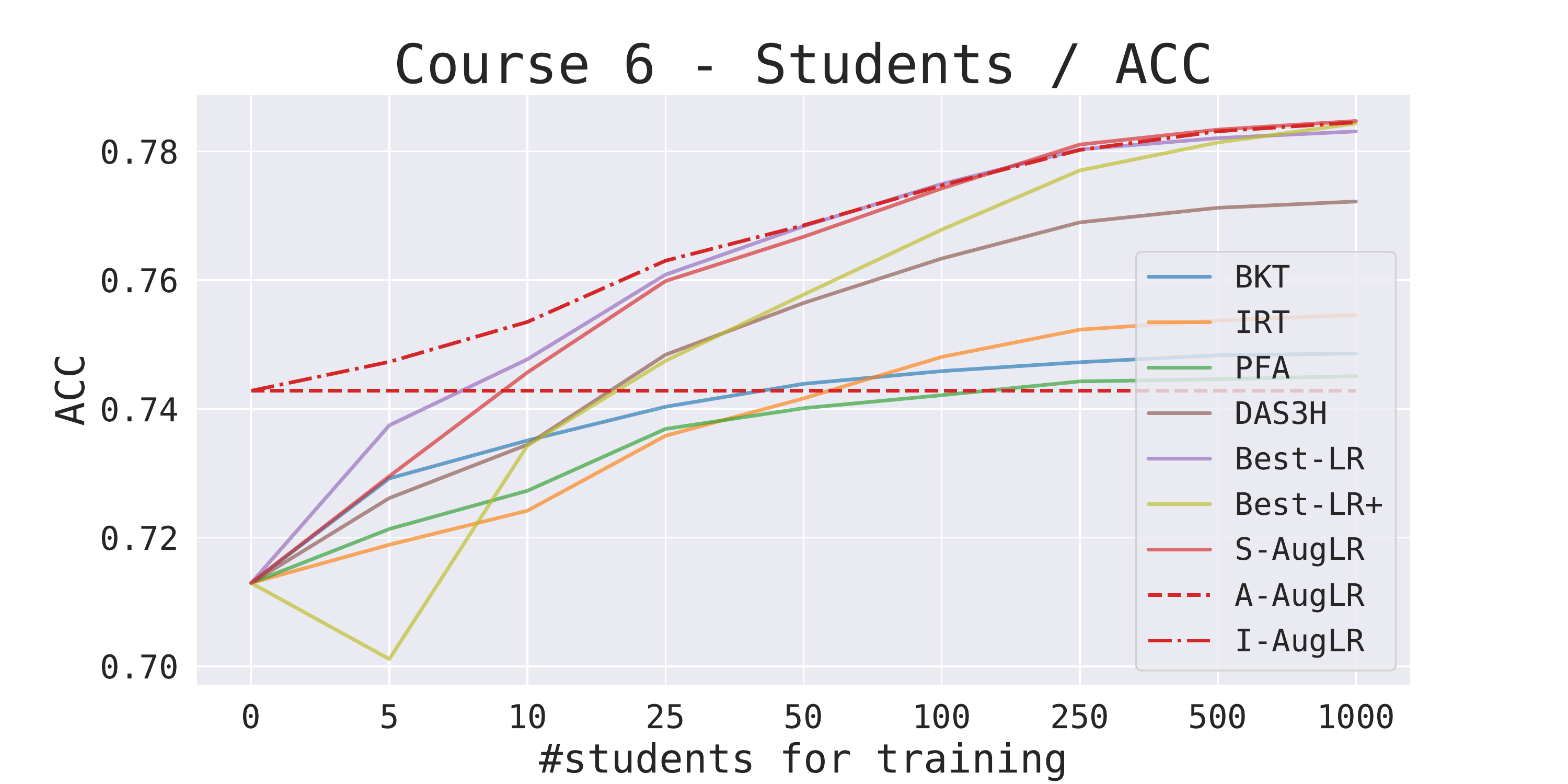}
    \includegraphics[width=0.435\textwidth]{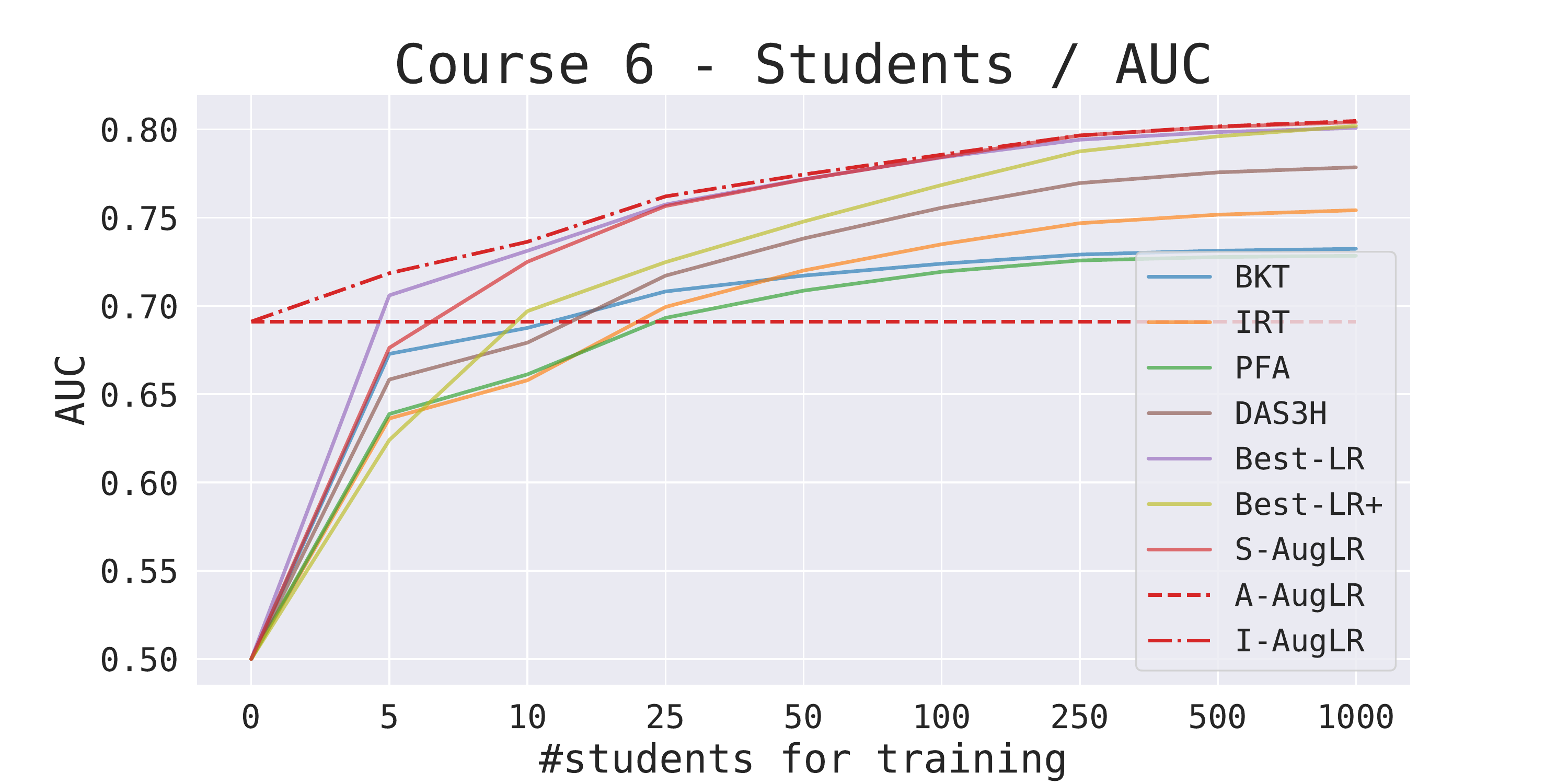}
    \includegraphics[width=0.435\textwidth]{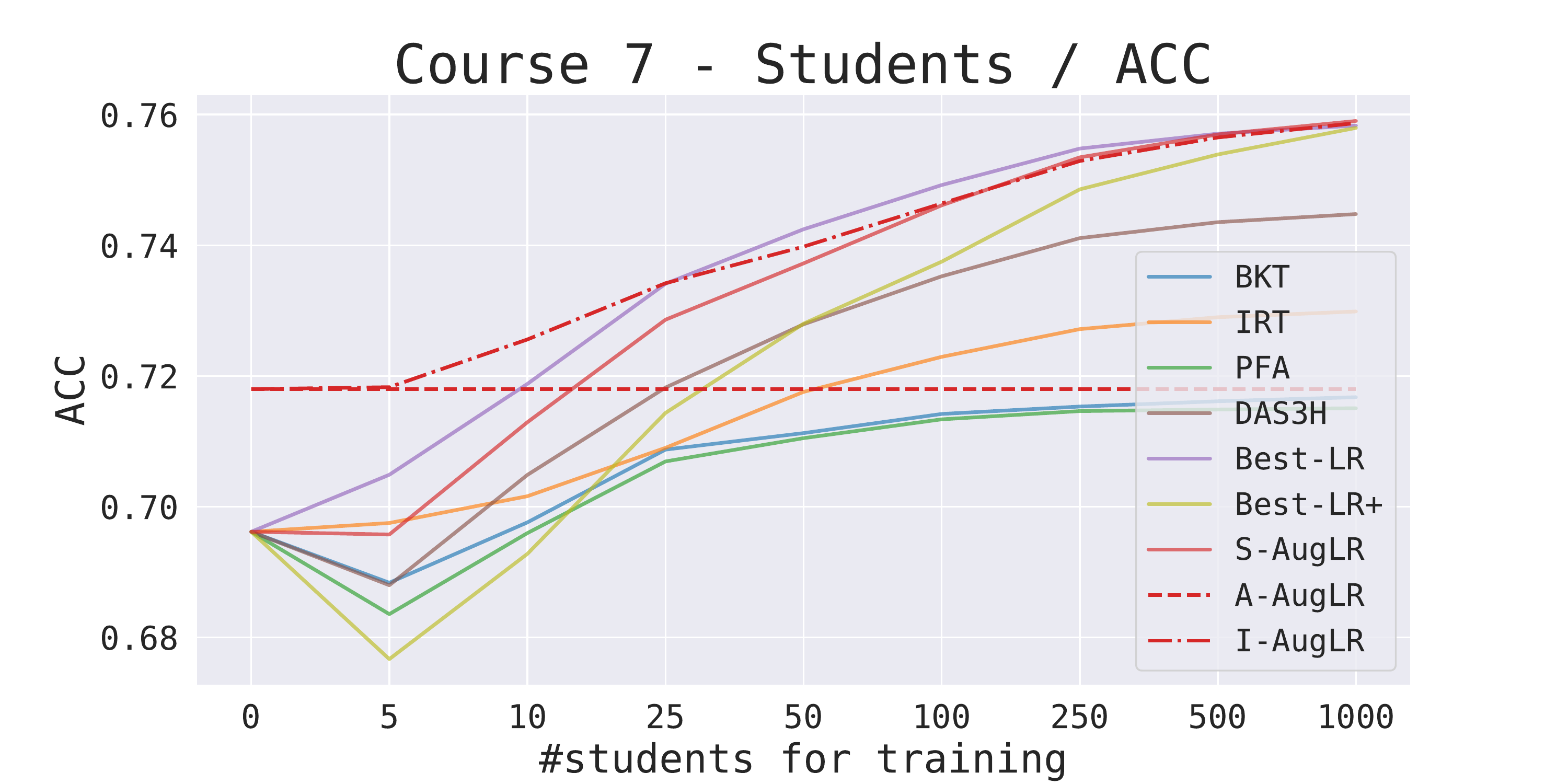}
    \includegraphics[width=0.4255\textwidth]{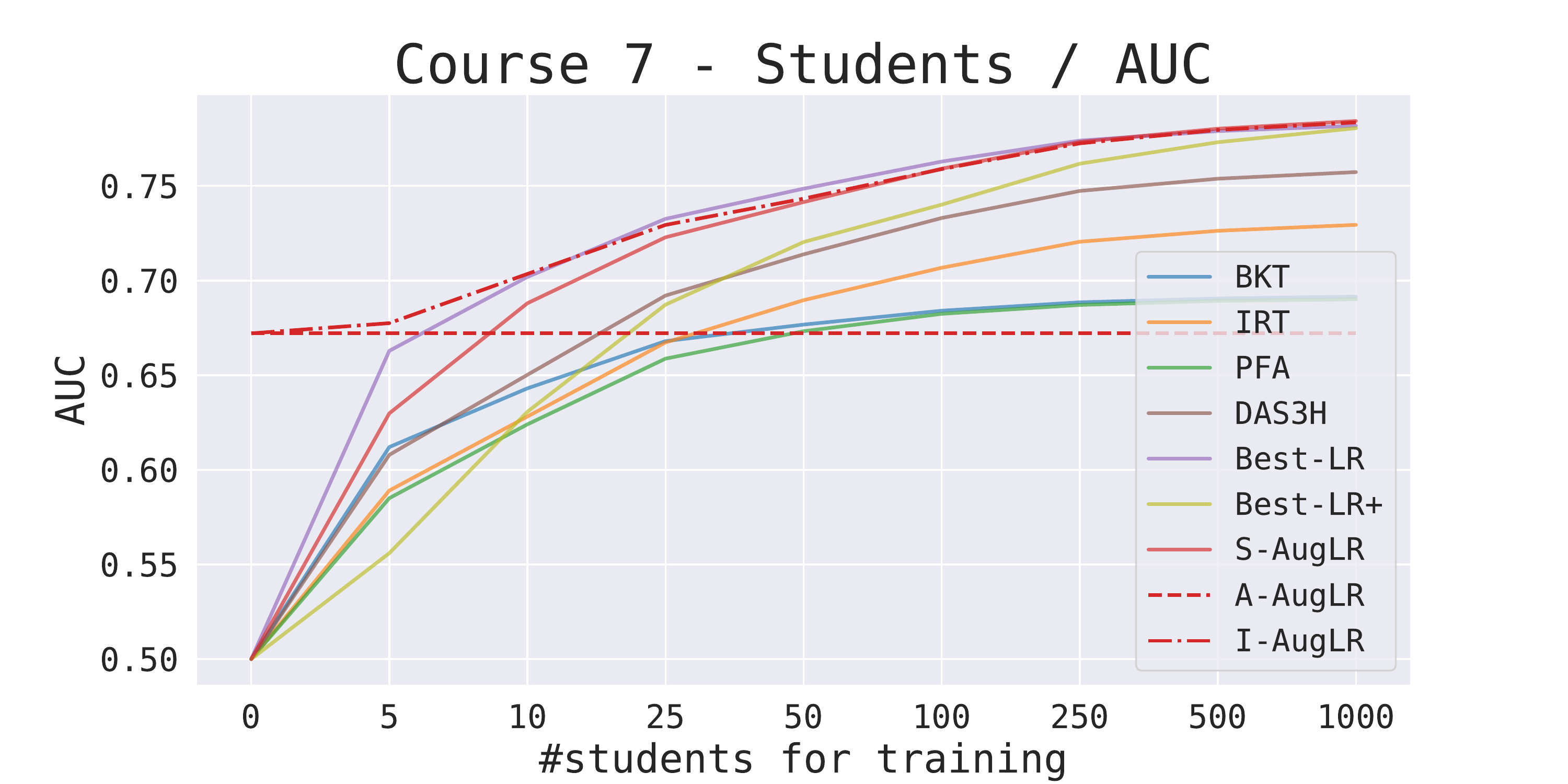}
    \includegraphics[width=0.4255\textwidth]{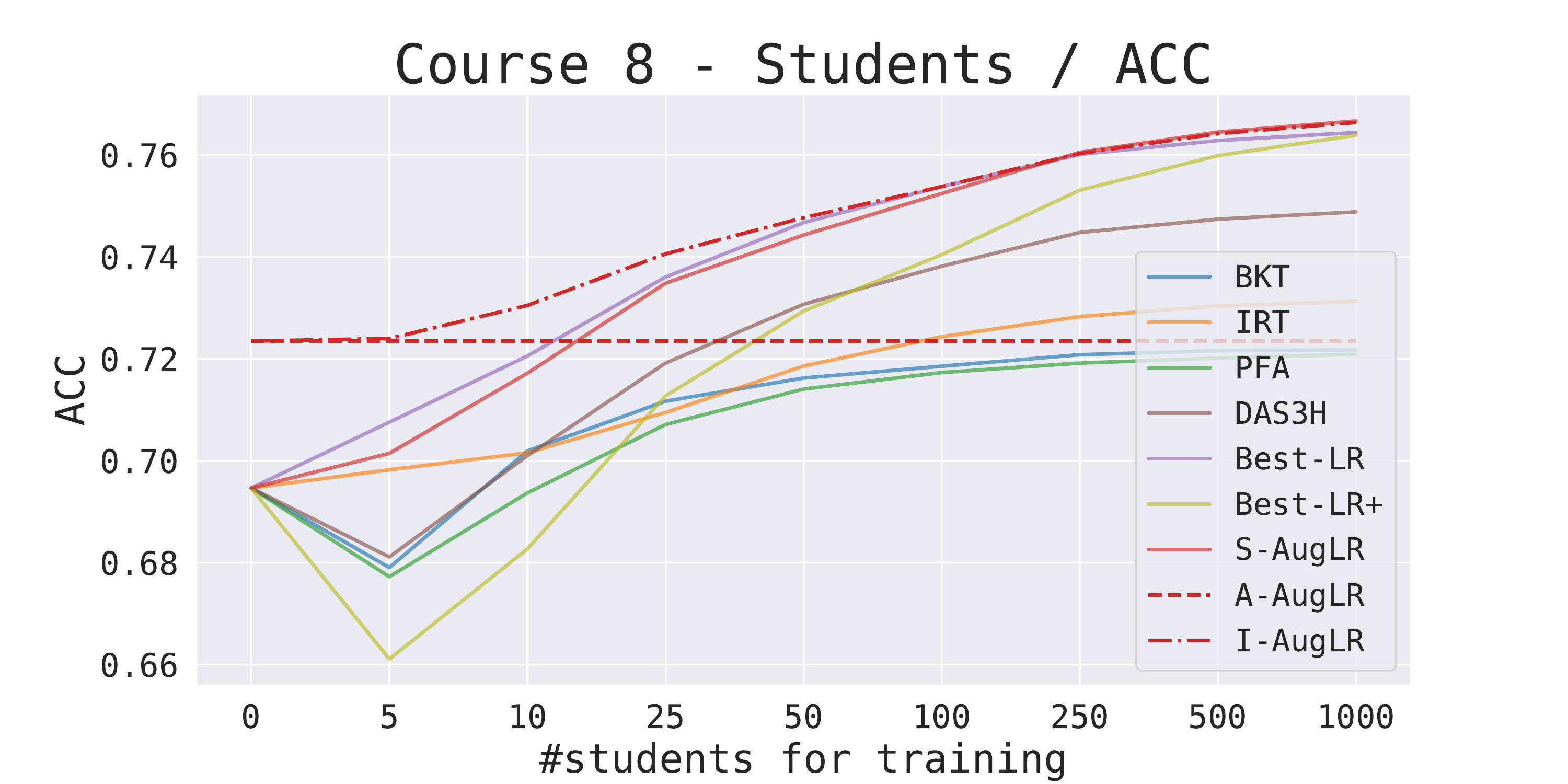}
    \includegraphics[width=0.4255\textwidth]{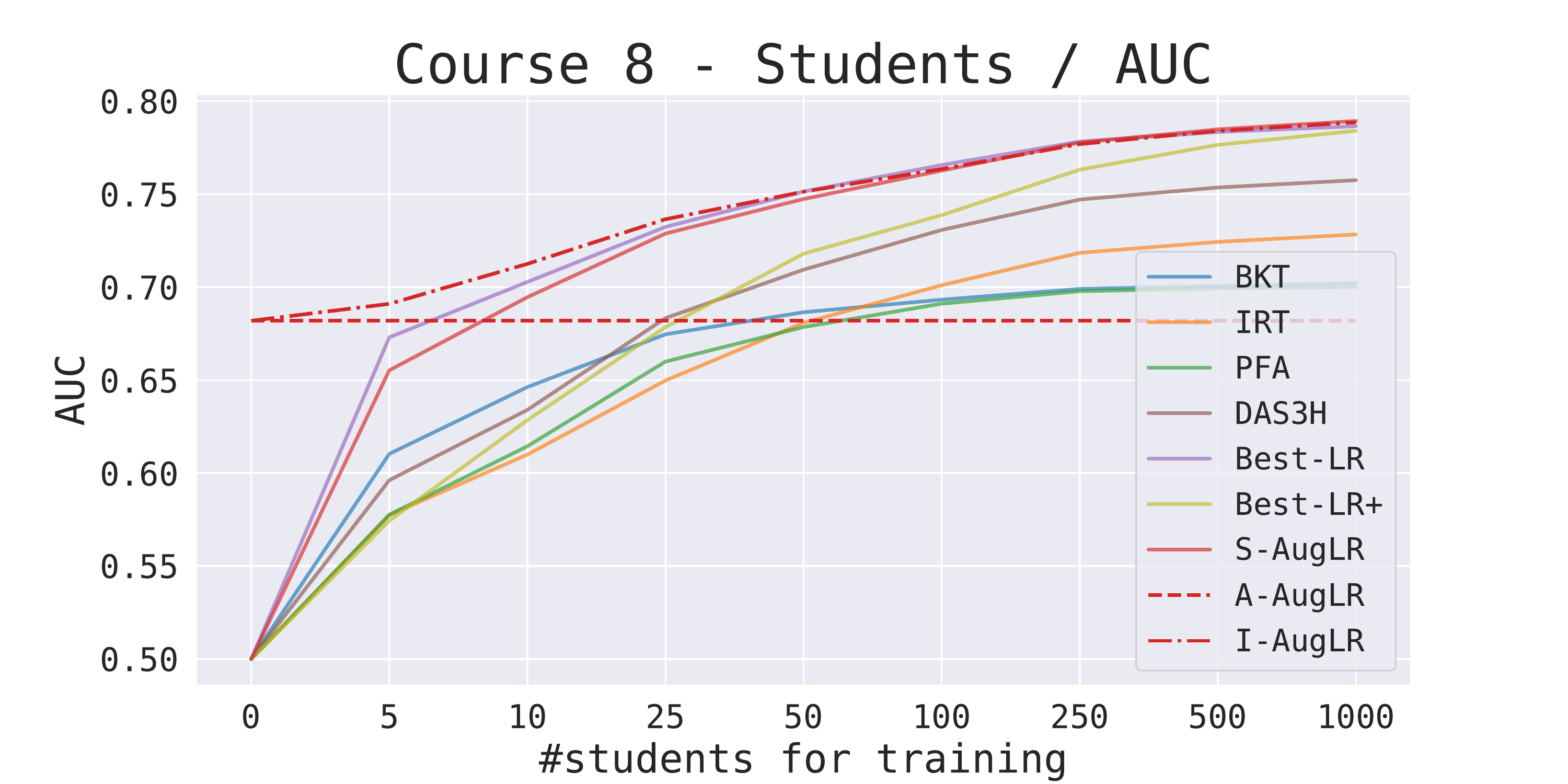}
    \includegraphics[width=0.4255\textwidth]{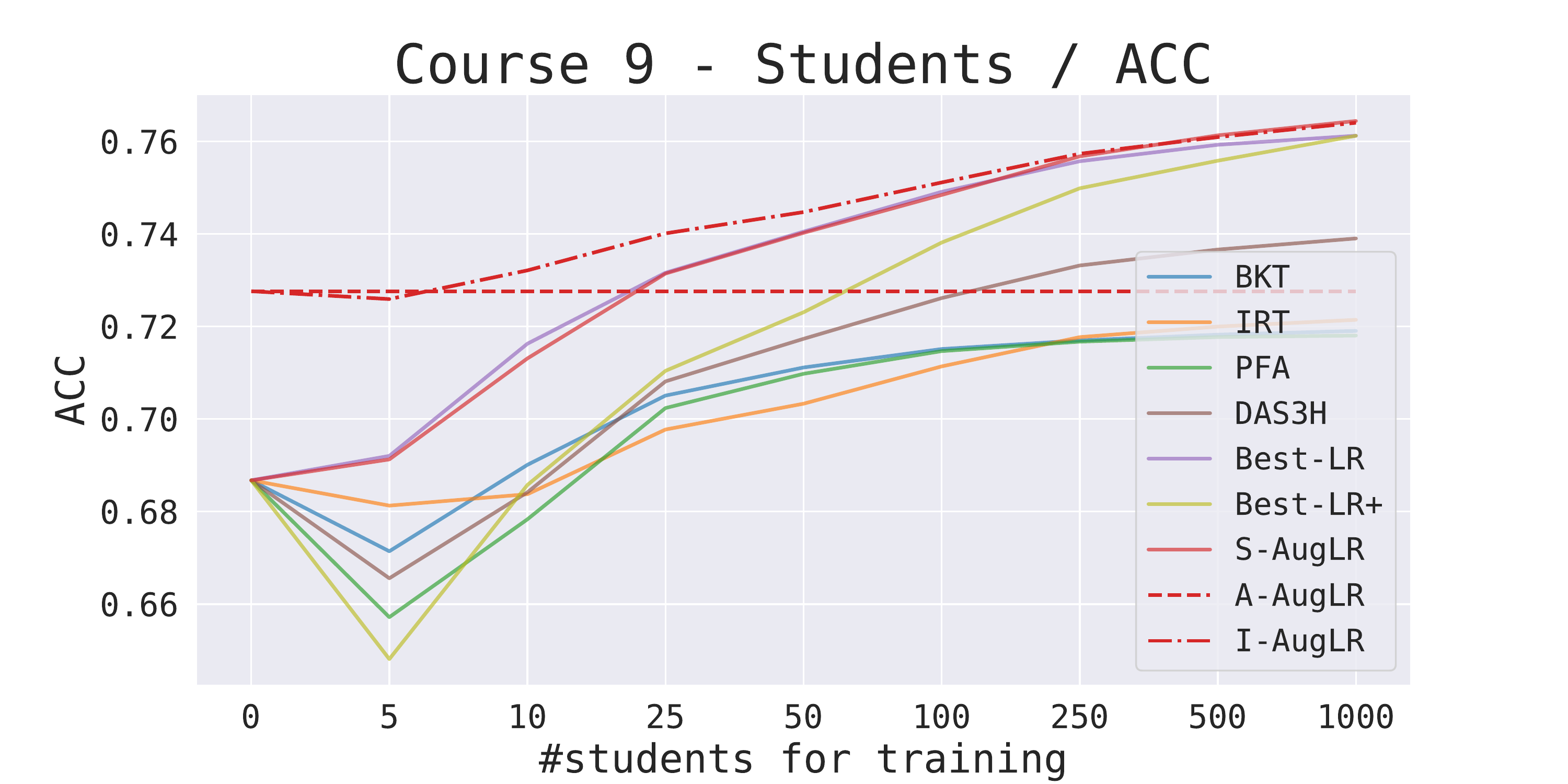}
    \includegraphics[width=0.4255\textwidth]{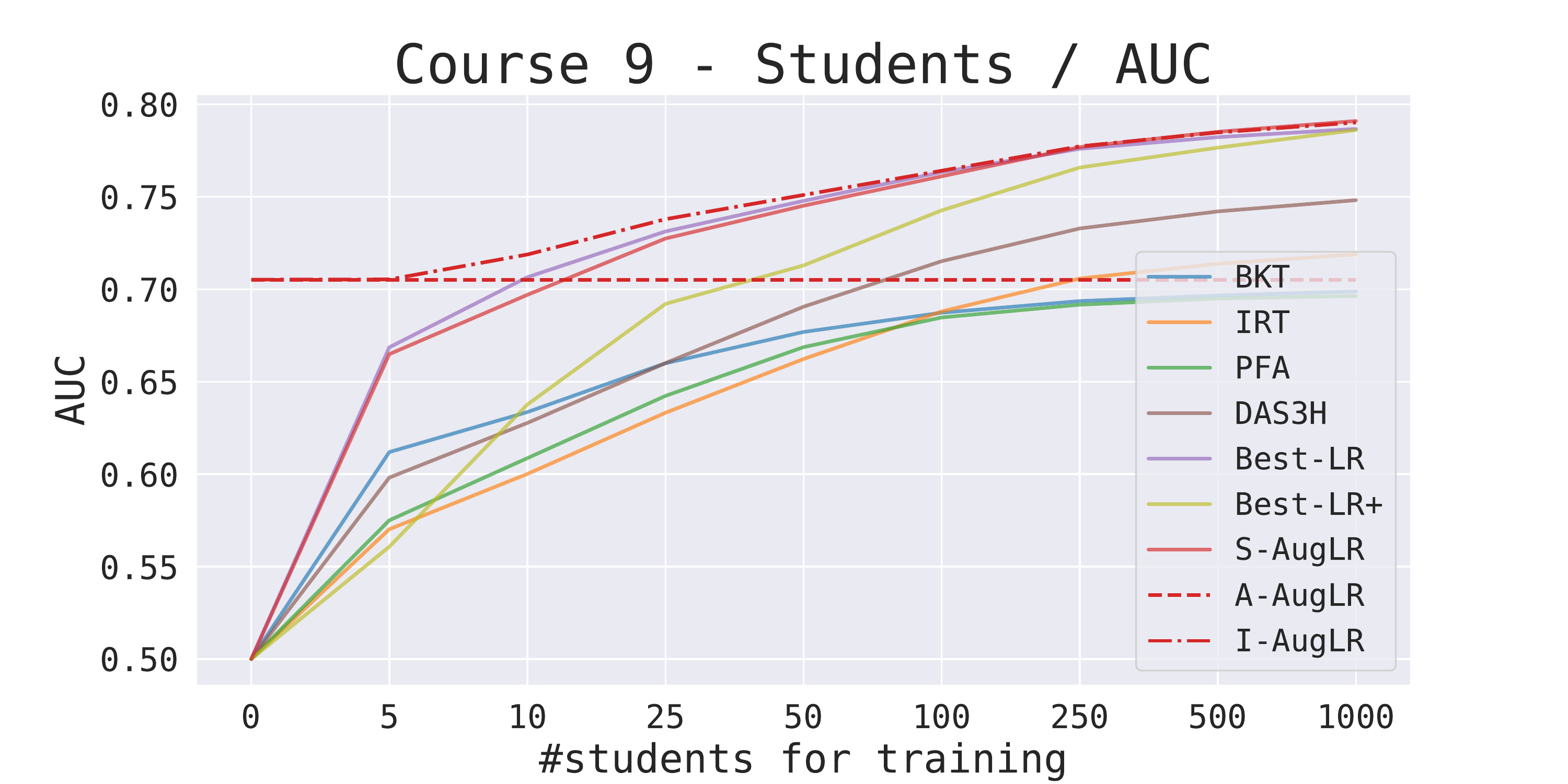}
    \includegraphics[width=0.4255\textwidth]{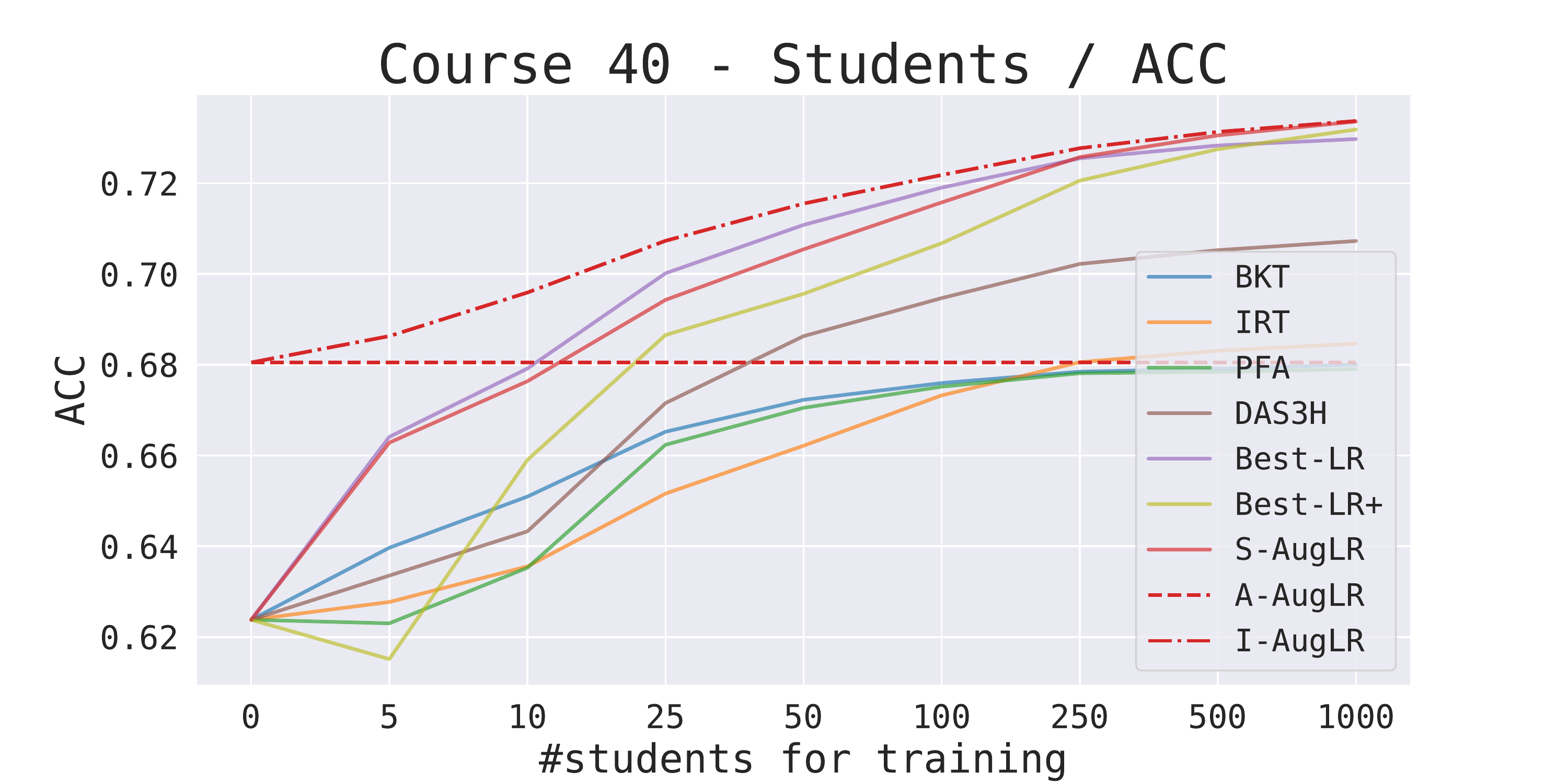}
    \includegraphics[width=0.4255\textwidth]{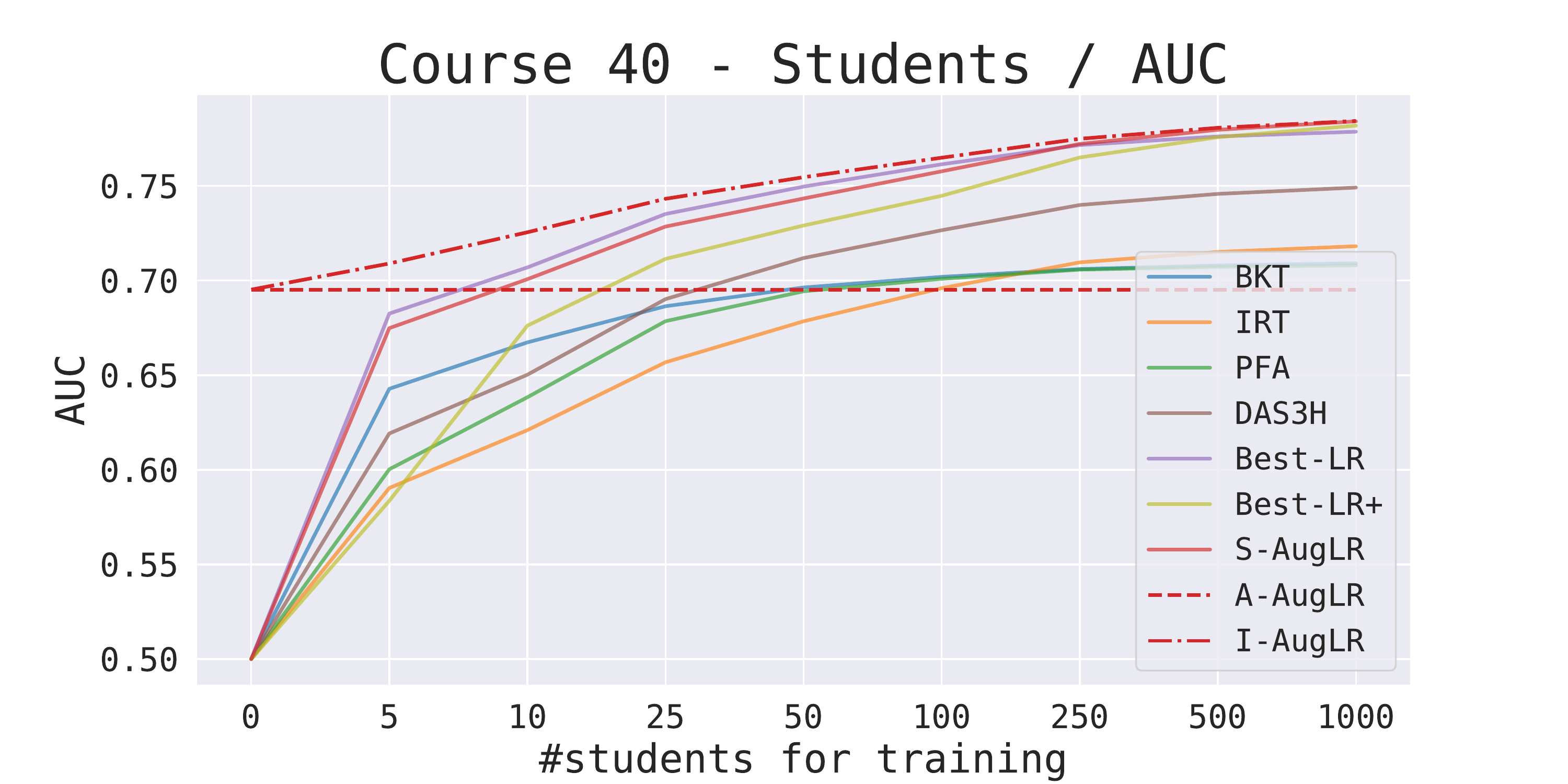}
    \caption{Relationship between the amount of available training data (measured in number of students) and ACC/AUC metrics achieved by the learned student performance models for each of the five courses. The dashed red line indicates the performance of a course-agnostic A-AugLR model that was pre-trained on student data from the other four courses and did not use any target course data. The dot-dashed red line indicates the performance of our inductive transfer approach (I-AugLR) which uses the additional data to tune the A-AugLR model to the target course. The course-specialized S-AugLR model is identical to I-AugLR, but does not leverage a pre-trained A-AugLR model. All results are averaged using a 5-fold cross validation.}
    \label{fig:inductive_transfer}
\end{figure*}

As discussed in Subsection~\ref{subsec:inductive_approach}, extending the course-agnostic A-AugLR feature set with features that capture question- and KC-difficulty parameters improves performance predictions substantially (Table~\ref{tab:advantage_item_skill}). Here, we evaluate our inductive transfer learning approach (I-AugLR) which uses small-scale target course data $D_T$ to tune a course-agnostic A-AugLR model -- pre-trained on log data from the other courses -- to the target course by learning course-specific difficulty parameters. We also evaluate the performance of a course-specific model (S-AugLR) which use the same feature set as I-AugLR, but does not leverage a pre-trained A-AugLR model. We measure the amount of target course data used for tuning in number of students who reached the end of the course. We experiment with student numbers in $\{0, 5, 10, 25, 50, 100, 250, 500, 1000\}$. The 0 student case is equivalent to the naive transfer setting.

Figure~\ref{fig:inductive_transfer} compares the performance of our inductive transfer learning method (I-AugLR) with conventional student performance modeling approaches and S-AugLR approach which were trained using only target course data $D_T$. By tuning a pre-trained A-AugLR model, I-AugLR is able to mitigate the cold-start problem for all courses and benefits from small-scale log data. Given as little as data from 10 students, the I-AugLR models consistently outperform standard BKT and PFA models that were trained on log data from thousands of target course students (Table~\ref{tab:traditional}). Among all considered performance models, I-AugLR yields the most accurate performance prediction up to 25 students for \texttt{C7}, up to 100 students for \texttt{C6} and \texttt{C8} and up to 250 students for \texttt{C9} and \texttt{C40}. Among the non-transfer learning approaches, Best-LR is most data efficient and yields the best performance predictions when training on up to 500 students. Best-LR+ builds upon Best-LR by using various additional features. While Best-LR+ outperforms Best-LR when training on thousands of students (Table~\ref{tab:traditional}), it performs worse when only limited training data is available.

\section{Discussion}
\label{sec:discussion}

Our experiments have shown that the proposed transfer learning techniques are able to mitigate the student performance modeling cold-start problem for new courses by leveraging student interaction data from existing courses. In the naive transfer setting where no target course data is available, the course-agnostic A-AugLR models that were trained on log data from existing courses yielded prediction accuracy on par with standard BKT and PFA models that use training data from thousands of students in the new course. One key ingredient of our course-agnostic models is additional information about question difficulty and learning context provided by human domain experts during content creation. While these features improve performance predictions, the need for manual annotations puts an additional load on the content creators. As an alternative to manual annotations, multiple techniques for inferring question difficulty estimates directly from the question text have been proposed~\cite{Benedetto2020:R2DE, Benedetto2020:Introducing, Loginova2021:Towards, Huang2021:Disenqnet}. Count features derived from manually specified KC pre-requisite graphs are beneficial when training and testing performance models on data from the same course~\cite{Schmucker2021:Assessing}, but they did not improve the course-agnostic performance predictions in our naive transfer setting. KC prerequisite information is highly domain specific and simple count features are not enough for knowledge transfer between courses. Future work on learning higher-order prerequisite graph features using deep learning techniques~\cite{Zhou2020:Graph} might improve transfer performance.

In order to be applicable to any course, the naive transfer learning models discussed in this paper avoid features that capture course-specific attributes. One inherent limitation of the naive transfer learning setting is that our models cannot learn parameters related to the difficulty of individual questions and KCs in the target course and only have access to discrete difficulty labels assigned by human domain experts. Table~\ref{tab:advantage_item_skill} trains and tests on target-course data to compare the performance of \emph{course-agnostic} A-AugLR models with A-AugLR models that learn additional \emph{course-specific} question- and KC- difficulty parameters. On average across the five courses, the A-AugLR models which were allowed to learn course-specific parameters yielded $4.5\%$ higher ACC and $7.9\%$ higher AUC scores than their course-agnostic counterparts. This emphasizes the importance of learning difficulty parameters from log data in addition to the difficulty ratings provided by human experts. Future work might compare the utility of difficulty ratings provided by domain experts with difficulty ratings extracted from the question text~\cite{Benedetto2020:R2DE, Benedetto2020:Introducing, Loginova2021:Towards, Huang2021:Disenqnet}. 

In the inductive transfer setting we use small-scale target course data (e.g., collected during a pilot study) to tune pre-trained course-agnostic student performance models. This allows us to overcome the limitations of the naive transfer setting by learning target course specific question- and KC- difficulty parameters. Our parameter regularized transfer approach yields better performance predictions than conventional modeling approaches when only limited target course data (<100 students) is available (Subsection~\ref{subsec:inductive_experiments}). This makes the inductive transfer approach an attractive option when one can run a pilot study with a small number of students before large-scale deployment. Surprisingly, we found that among the non-transfer learning approaches, Best-LR~\cite{Gervet2020:Deep} yielded the most accurate predictions when training on less than 500 students for all five courses. This is interesting because low-parametric models such as BKT~\cite{Corbett1994:Knowledge}, PFA~\cite{Pavlik2009:Performance} or IRT~\cite{Rasch1993:Probabilistic} are commonly believed to be more data efficient than more complex logistic regression models that contain many more parameters that must be learned. What sets Best-LR apart from these three models, is that its parameters describe student performance using multiple levels of abstraction (question-, KC- and overall-level). Future work might investigate this phenomenon further using log data from multiple ITSs. 

One limitation of our study is that it focuses on a set of five courses offered by the same ITS. This has advantages because the course log data is of consistent format and content creators follow similar protocols. Still, it prevents us from answering the question of whether student performance models are transferable between different tutoring systems~\cite{Baker2019:Challenges}. Another related limitation is that all considered courses cover mathematics topics for elementary school students. Our study did not investigate the transferability of student performance models across different subjects or grade levels (i.e, middle school, high school, \dots). While our experiments indicate that our naive transfer approach is robust towards the choice of source and target course pairing (Subsection~\ref{subsec:naive_experiments}), results may vary in settings with larger differences between individual courses.

\section{Conclusion}
\label{sec:conclusion}

The increasing popularity of online tutoring systems induces a need for student performance modeling techniques that are flexible enough to support frequent new course releases as well as changes to existing courses. Relatedly, this paper proposed two transfer learning approaches for mitigating the cold-start problem that arises when a new (target) course is introduced for which no training data is available. In the naive transfer setting where no target course data is available, we rely on student interaction sequences from existing courses to learn course-agnostic student performance models that can be applied to any future course. Through the inclusion of question difficulty and learning context information provided by human domain experts during content creation, our course-agnostic models enable performance prediction accuracy on par with conventional BKT and PFA models that were trained on target course data. In the inductive transfer setting where small-scale target course data is available (e.g., collected during a pilot study), we show how one can tune pre-trained course-agnostic models to a specific target course by learning question- and KC- difficulty parameters. Our experimental evaluation on student data from five different mathematics courses showed how both transfer approaches mitigate the cold-start problem successfully. This work represents a first step in the design of student performance models that are transferable between different ITS courses and forms a base for effective adaptive instruction for early adopter students.

\bibliographystyle{ACM-Reference-Format}
\bibliography{bibliography}

\end{document}